\documentclass[10pt,twocolumn,letterpaper]{article}

\usepackage{cvpr}              

\usepackage{adjustbox}
\usepackage{graphicx}
\usepackage{multirow}
\usepackage{tabularx}
\usepackage{array}
\usepackage{makecell}
\usepackage{arydshln}
\usepackage{float}
\usepackage{tikz}
\usepackage{subcaption}
\usepackage{colortbl}
\usepackage{xcolor}
\usetikzlibrary{fit}
\usepackage{enumitem}

\definecolor{cvprblue}{rgb}{0.21,0.49,0.74}
\usepackage[pagebackref,breaklinks,colorlinks,allcolors=cvprblue]{hyperref}

\def\paperID{12} 
\def\confName{CVPR}
\def\confYear{2025}

\title{AGILE: A Diffusion-Based Attention-Guided Image and Label Translation for Efficient Cross-Domain Plant Trait Identification}

\author{Earl Ranario, Lars Lundqvist, Heesup Yun, Brian N. Bailey, J. Mason Earles \\
University of California, Davis\\
{\tt\small \{ewranario, llund, hspyun, bnbailey, jmearles\}@ucdavis.edu}
}

\begin{document}
\maketitle
\begin{abstract}

Semantically consistent cross-domain image translation facilitates the generation of training data by transferring labels across different domains, making it particularly useful for plant trait identification in agriculture. However, existing generative models struggle to maintain object-level accuracy when translating images between domains, especially when domain gaps are significant. In this work, we introduce AGILE (\textbf{A}ttention-\textbf{G}uided \textbf{I}mage and \textbf{L}abel Translation for \textbf{E}fficient Cross-Domain Plant Trait Identification)\footnote{\url{https://github.com/plant-ai-biophysics-lab/AGILE}}, a diffusion-based framework that leverages optimized text embeddings and attention guidance to semantically constrain image translation. AGILE utilizes pretrained diffusion models and publicly available agricultural datasets to improve the fidelity of translated images while preserving critical object semantics. Our approach optimizes text embeddings to strengthen the correspondence between source and target images and guides attention maps during the denoising process to control object placement. We evaluate AGILE on cross-domain plant datasets and demonstrate its effectiveness in generating semantically accurate translated images. Quantitative experiments show that AGILE enhances object detection performance in the target domain while maintaining realism and consistency. Compared to prior image translation methods, AGILE achieves superior semantic alignment, particularly in challenging cases where objects vary significantly or domain gaps are substantial.

\end{abstract}    
\section{Introduction}
\label{sec:intro}

\begin{figure}[t]
    \centering
    \includegraphics[width=1.0\linewidth]{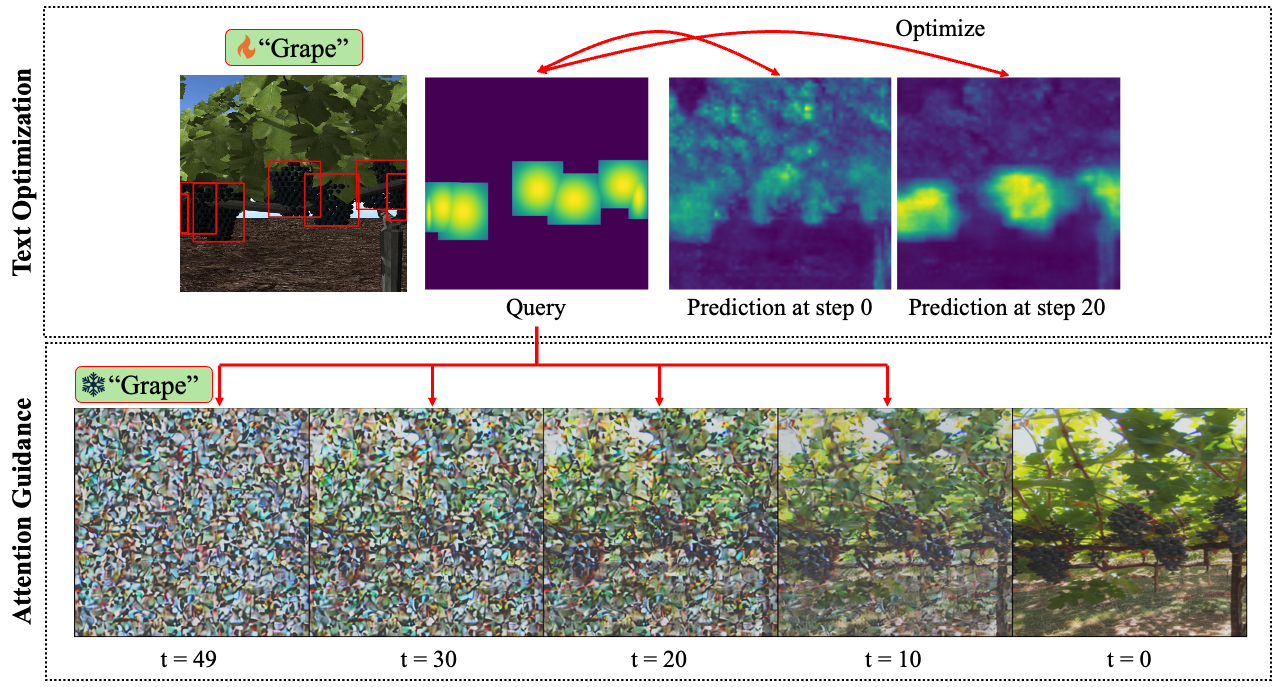}
    \caption{We use labels from synthetic data to gain object domain knowledge represented by text labels by optimizing for semantic correspondences between the source and target domains.}
    \label{fig:semantic}
\end{figure}

\begin{figure*}[h]
    \centering
    \includegraphics[width=1.0\linewidth]{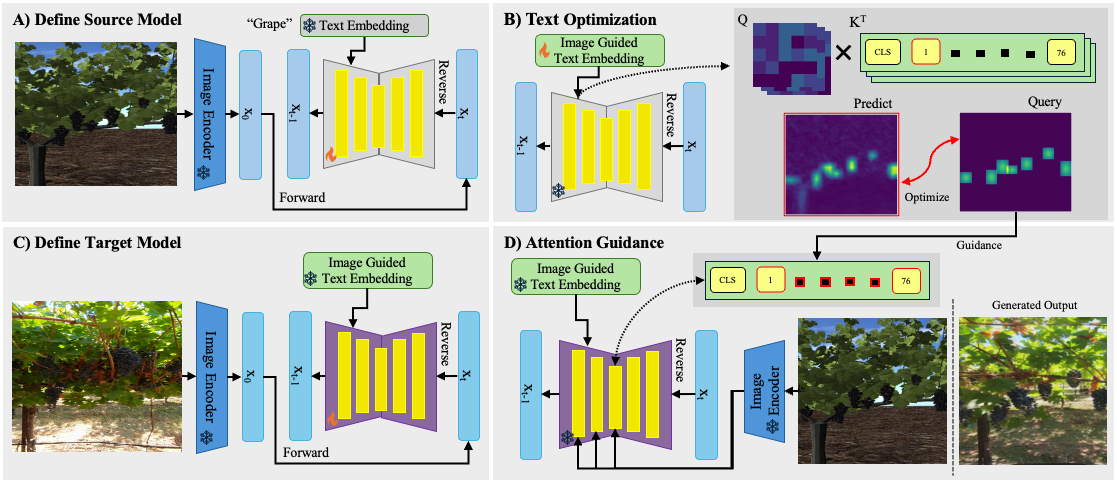}
    \caption{AGILE uses pretrained diffusion models to find semantic correspondences between unpaired source and target images. We optimize text embeddings through query attention maps generated from labeled source images, guiding the model to focus on desired regions in the target domain. Attention guidance is applied during the denoising process to enhance control over semantic alignment, achieving improved consistency in translation between source and target domains.}
    \label{fig:method}
\end{figure*}

\subsection{Computer Vision in Agriculture}
Computer vision is used in a wide range of agricultural tasks such as plant phenotyping, disease detection, and yield estimation. Such tasks involve the identification of plant-specific traits or characteristics that reflect the condition of the plant. This allows farmers or plant breeders to identify key phenotypic traits in plants that improve decision-making. For example, Palacios et al. \cite{palacios_early_2023} implemented a segmentation model to detect visible berries and canopy features to predict the yields of different varieties of grapes. Palacios further states that their results would improve should a higher number of diverse data points be used to build the models. Chen et al. \cite{chen_detection_2020} detected rice plant diseases based on deep transfer learning, utilizing pre-trained models which was trained on a large amount of images and fine-tuned on a smaller, domain specific dataset.

Although machine learning tools have allowed for high-throughput identification of plant traits, the performance of these tools is limited by the availability of labeled data and resource constraints \cite{jiang2020}. There is an emphasis on expanding public image datasets for agricultural tasks, but this alone may not adequately address the complexities of multi-domain scenarios \cite{lu2020}. For instance, a model trained on one domain may not generalize well to another domain due to differences in lighting, camera angle, or plant species.
The general approach is to label new data tailored to specific domains, but this process is costly and time-consuming \cite{li_label-efficient_2023}. However, recent advances in artificial intelligence (AI) provide new capabilities in improving the efficiency of labeling.

\subsection{Domain Translation}

Some of the many applications of generative AI to improve labeling efficiency include image-to-image translation, text-to-image generation, and style transfer. These methods have been applied to a variety of applications, including medical imaging, autonomous driving, and video games. For the case of image-to-image translation, by using existing labeled public datasets, it is possible to translate the images to another domain while maintaining the semantics of the labeled object, where semantics refers to the structural features of an object, such as its shape, color, and positional relationships within the image. 

Generative Adversarial Networks (GANs) \cite{goodfellow2014generativeadversarialnetworks, zhu2020unpairedimagetoimagetranslationusing} is an early example of image-to-image translation for unpaired images. More recently, Stable Diffusion models \cite{rombach2022highresolutionimagesynthesislatent, podell2023sdxlimprovinglatentdiffusion} have enabled high-quality image generation through various conditioning mechanisms. Although diffusion models such as DALL-E \cite{ramesh_zero-shot_2021,ramesh2022hierarchicaltextconditionalimagegeneration,betker_improving_nodate} and Stable Diffusion are capable of generating complex scenes, using them for semantically constrained image-to-image translation comes with challenges. First, for most diffusion-based models, training a model that can translate an image from one domain to another requires paired images, which is difficult to obtain. Second, images collected in the field do not come with text descriptions. For instance, the prompt ``grapes in a vineyard'' could output different types of grapes in various vineyard settings. Third, the generated images may not be semantically accurate or not contain the desired object. For the case of image-to-image translation, the user may want to keep an object in a specific location based on the semantics of the input image.

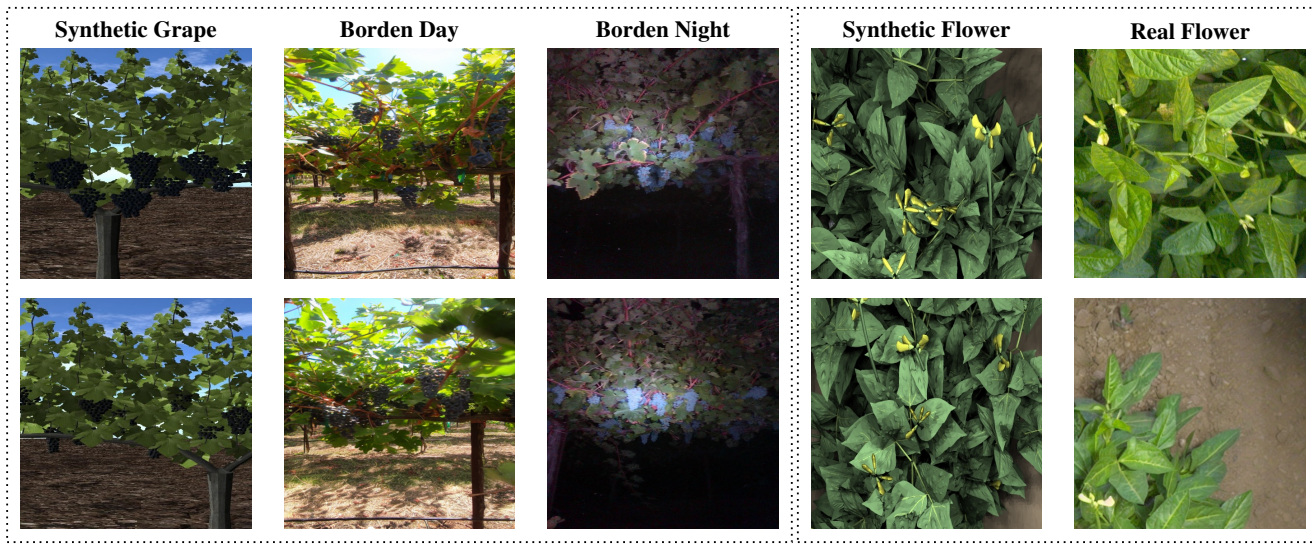
\begin{figure*}[t]
    \centering
    \begin{adjustbox}{width=\textwidth}
        \begin{tikzpicture}
            \node (title1) at (0,2) {\textbf{Synthetic Grape}};
            \node (title2) at (4,2) {\textbf{Borden Day}};
            \node (title3) at (8,2) {\textbf{Borden Night}};
            \node (title4) at (12,2) {\textbf{Synthetic Flower}};
            \node (title5) at (16,2) {\textbf{Real Flower}};

            \node (img1) at (0,0) {\includegraphics[width=3.5cm]{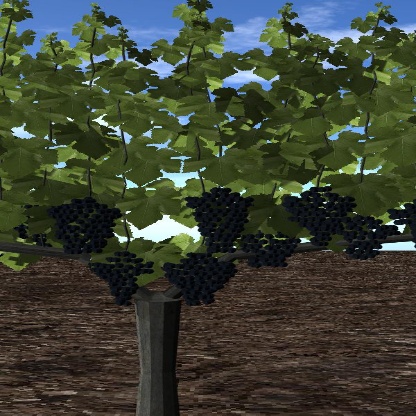}};
            \node (img2) at (4,0) {\includegraphics[width=3.5cm]{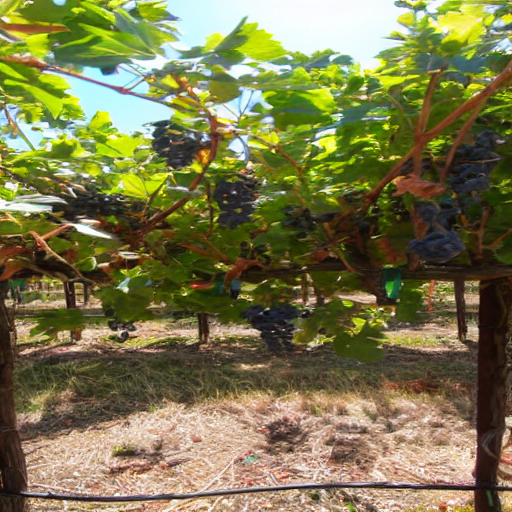}};
            \node (img3) at (8,0) {\includegraphics[width=3.5cm]{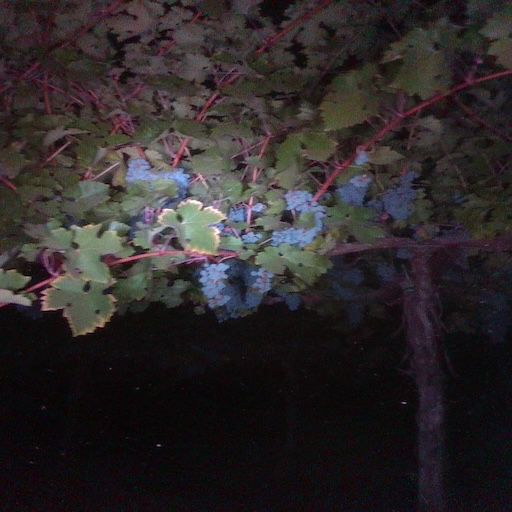}};
            \node (img4) at (12,0) {\includegraphics[width=3.5cm]{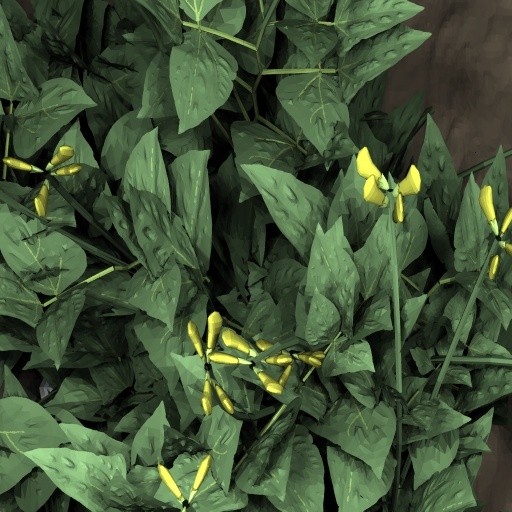}};
            \node (img5) at (16,0) {\includegraphics[width=3.5cm]{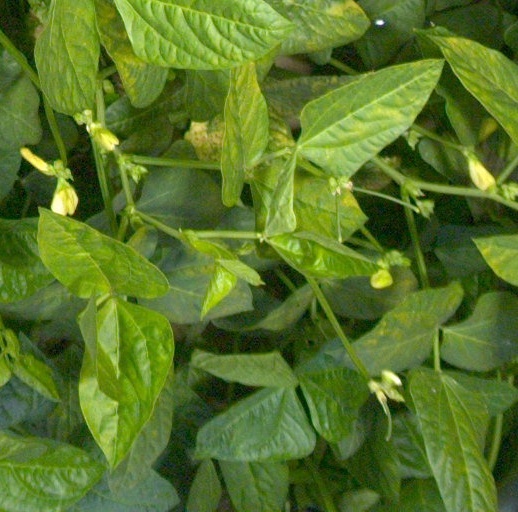}};

            \node (img6) at (0,-3.8) {\includegraphics[width=3.5cm]{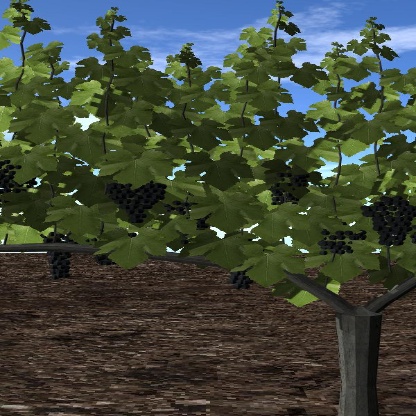}};
            \node (img7) at (4,-3.8) {\includegraphics[width=3.5cm]{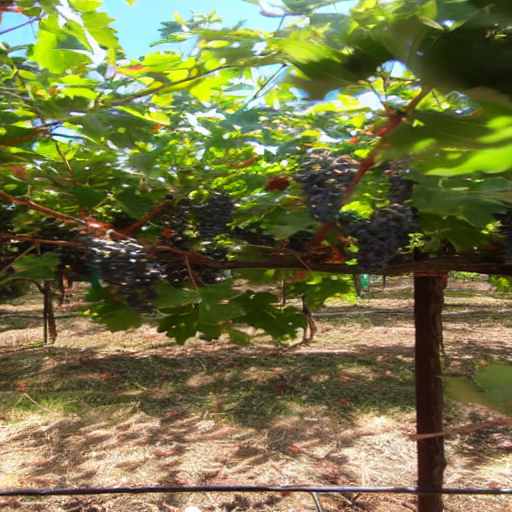}};
            \node (img8) at (8,-3.8) {\includegraphics[width=3.5cm]{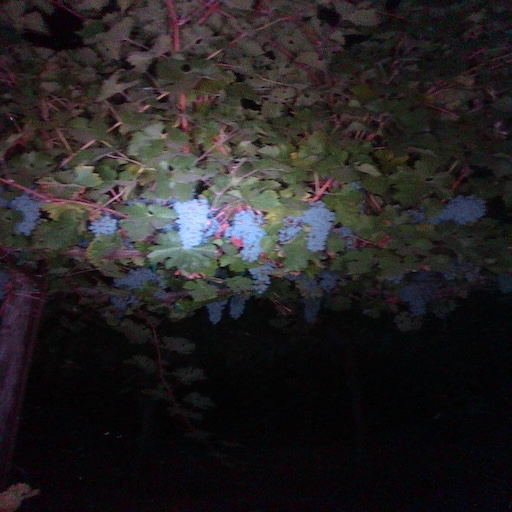}};
            \node (img9) at (12,-3.8) {\includegraphics[width=3.5cm]{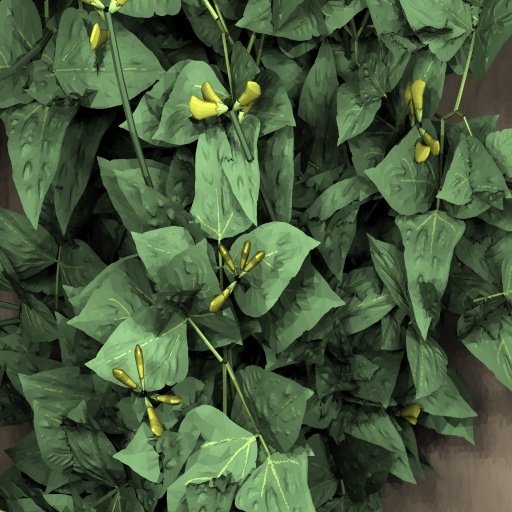}};
            \node (img10) at (16,-3.8) {\includegraphics[width=3.5cm]{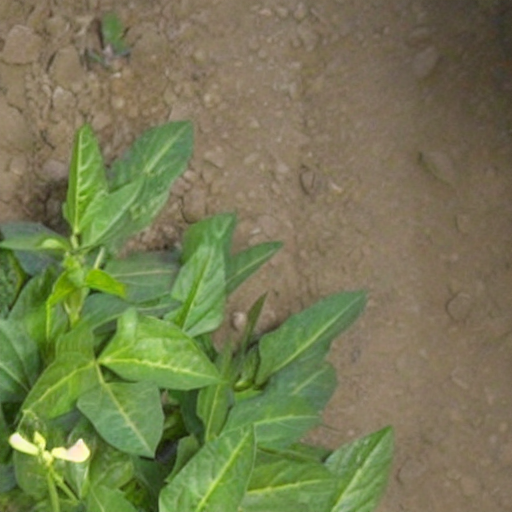}};

            \node[draw, dotted, thick, inner sep=0.8mm, fit=(title1) (title2) (title3) (img6) (img7) (img8)] {};

            \node[draw, dotted, thick, inner sep=0.8mm, fit=(title4) (title5) (img9) (img10)] {};

        \end{tikzpicture}
    \end{adjustbox}

    \caption{The dataset is pulled from AgML, a machine learning library for agricultural datasets. The original synthetic images were generated from Helios, a 3D Plant and Environment Biophysical Modeling Framework \cite{bailey_helios_2019, lei_simulation_2024}. Synthetic images is treated as the source domain for its capability to generate an infinite amount of labeled images. We train and evaluate our method on object detection tasks and constrain the translation within the same plant.}
    \label{fig:datasets}
\end{figure*}

Therefore, this paper studies how to efficiently utilize existing labeled images to improve semantic accuracy in image-to-image translation tasks, specifically for plant trait identification. We propose a diffusion-based method \textbf{A}ttention-\textbf{G}uided \textbf{I}mage and \textbf{L}abel Translation for \textbf{E}fficient Cross-Domain Plant Trait Identification (AGILE), which utilizes existing pretrained diffusion models and public agricultural datasets to generate labeled images for specific domains. The idea is to use labels generated for supervised tasks to gain object domain knowledge represented by text labels. Through this, we should be able to find semantic correspondences between the source and target domain and guide it to desired regions. Our contributions include:
\setlength{\parskip}{5pt}
\begin{itemize}
    \item Images collected in real-world settings often lack accompanying text descriptions and semantic alignment with text inputs. By optimizing prompt embeddings, we leverage existing labeled images to emphasize regions of interest, improving semantic knowledge. This allows us to have text-image correspondence with few labeled images.
    \item With semantically-aware, optimized prompt embeddings, we can control object semantics in the target domain through attention guidance during the de-noising process of a diffusion-based model. This allows labels to be transferrable from the source to target domains.
    \item We compare the performance between the source, target and generated data for object detection tasks.
\end{itemize}
\section{Related Work}
\label{sec:related_work}

\subsection{Image-to-Image Translation}

Image-to-image translation is the task of translating one possible representation of a scene into another, which was explored early with GANs \cite{isola2018imagetoimagetranslationconditionaladversarial}. Within the GAN framework, a generator creates an output image that aims to resemble a target image, while the discriminator evaluates how real or fake the output looks compared to the actual data. Fei et al. enlisted 3D crop models and GANs to semantically constrain fruit position and geometry \cite{fei_enlisting_2021}. CropGAN performs effectively when the source and target domains are similar, but its performance deteriorates significantly when applied to vastly different domains.

For the case of classifier-free diffusion-based models, conditional inputs can be included during the generation process, allowing for more creativity or control. However, diffusion models can dramatically change the content of the desired image and introduce unexpected changes in regions of interest. Parmar et al. developed a zero-shot image-to-image translation using the Stable Diffusion framework \cite{parmar2023zeroshotimagetoimagetranslation}. Their proposed method focuses on changing desired regions while keeping unrelated regions consistent by editing the text embedding space and using cross-attention guidance. While their approach effectively maintains the structure of the desired region, it struggles with handling complex features. Moreover, their method is limited to translating only a single object per image. Parmar et al. additionally addressed the slow processing speed and the reliance of paired data for model fine-tuning with CycleGan-Turbo and pix2pix-Turbo by adapting a single-step diffusion model to new domains through adversarial learning objectives \cite{parmar_one-step_2024}. Xu et al. introduced CycleNet, which incorporates cycle consistency into diffusion models to regularize image to image manipulation, without the need for paired data \cite{xu_cyclenet_2024}.

Zhang et al. developed a method, called ControlNet, to add conditional control for text-to-image diffusion models \cite{zhang2023addingconditionalcontroltexttoimage}. ControlNet locks the production ready large diffusion models and reuses their pretrained encoding layers to learn a diverse set of conditional controls. It also contains ``zero convolutions" (zero-initialized convolutional layers) that progressively grow the parameters from zero and ensure that no harmful noise could affect finetuning. Their method allows for various control methods to be used and is very fast to converge during training. However, as is the case for most diffusion-based methods, paired data needs to be accessible to streamline the training process.

\subsection{Semantic Correspondence}

Semantic correspondence is a core problem in computer vision tasks that relates to finding corresponding locations in images that are of the same semantic \cite{ma_image_2021,cho_cats_2023,gupta_asic_2023}. Hedin et al. leverages semantic knowledge within diffusion models to find locations in multiple images that have the same semantic meaning \cite{hedlin_unsupervised_2023}. Their key insight is that since recent diffusion models can generate photo-realistic images from text prompts only, there must be knowledge about semantic correspondences built-in within them. Therefore, they deduced that one may not need any ground-truth semantic correspondences between image pairs to find semantic correspondences. By exploiting the attention maps of latent diffusion models, one can identify the prompt corresponding to a particular image location. Given arbitrary input images, these attention maps should respond to the semantics of the prompt. They proposed a method, inspired by recent prompt-to-prompt text-based image editing \cite{mokady_null-text_2022}, to first optimize a randomly initialized text embedding to maximize the cross-attention score at a query location. Then, they find the semantically corresponding location in another image by using the pixel attaining the maximum attention map score within the target image. Attend and Excite \cite{chefer_attend-and-excite_2023} is an example of a similar approach. We can build upon this method by not relying on paired images by using the model's underlying knowledge of the object based on the given prompt.

\subsection{Cross-Attention Guidance}

Content preservation through cross-attention guidance involves maintaining the semantics of an image before and after diffusion translation by ensuring that the text-image cross-attention map remains consistent. Ma et al. demonstrated cross-attention guidance in their method, Directed Diffusion, which improves upon diffusion models by providing direct control over object placement within generated images \cite{ma_directed_2023}. Their approach uses text-image cross-attention maps to guide the positioning of labeled or specified objects while maintaining contextual coherence. However, the constraint is that text prompts provided must have text-image correspondence, which are not readily available in the field. Therefore, prior to attention guidance, we must first optimize text embeddings to find semantic correspondences between the text prompt and target images.
\section{Experimental Setup}
\label{sec:exp_setup}

\begin{figure}[t]
    \centering
    \includegraphics[width=1.0\linewidth]{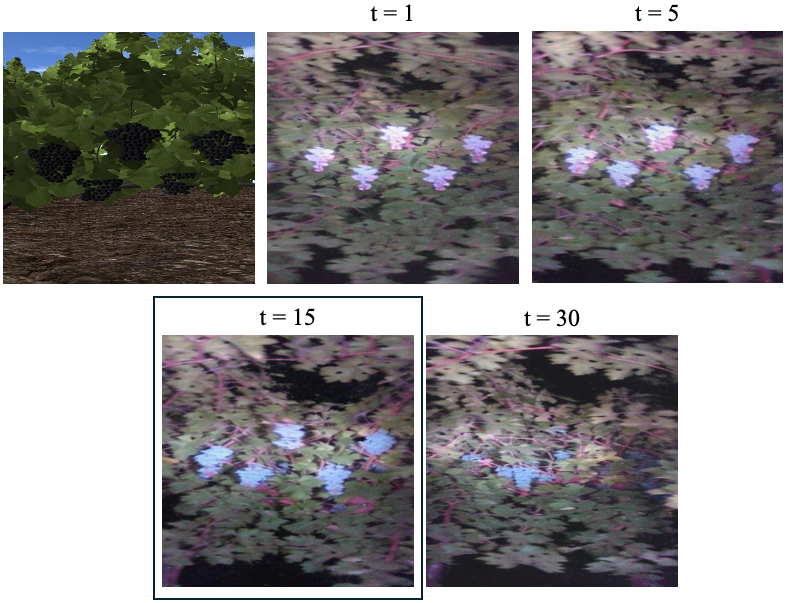}
    \caption{The displayed timesteps indicate when attention guidance is halted. The optimal stopping range is between $t=5$ to $t=15$, as this preserves object structure and color effectively.}
    \label{fig:timesteps}
\end{figure}

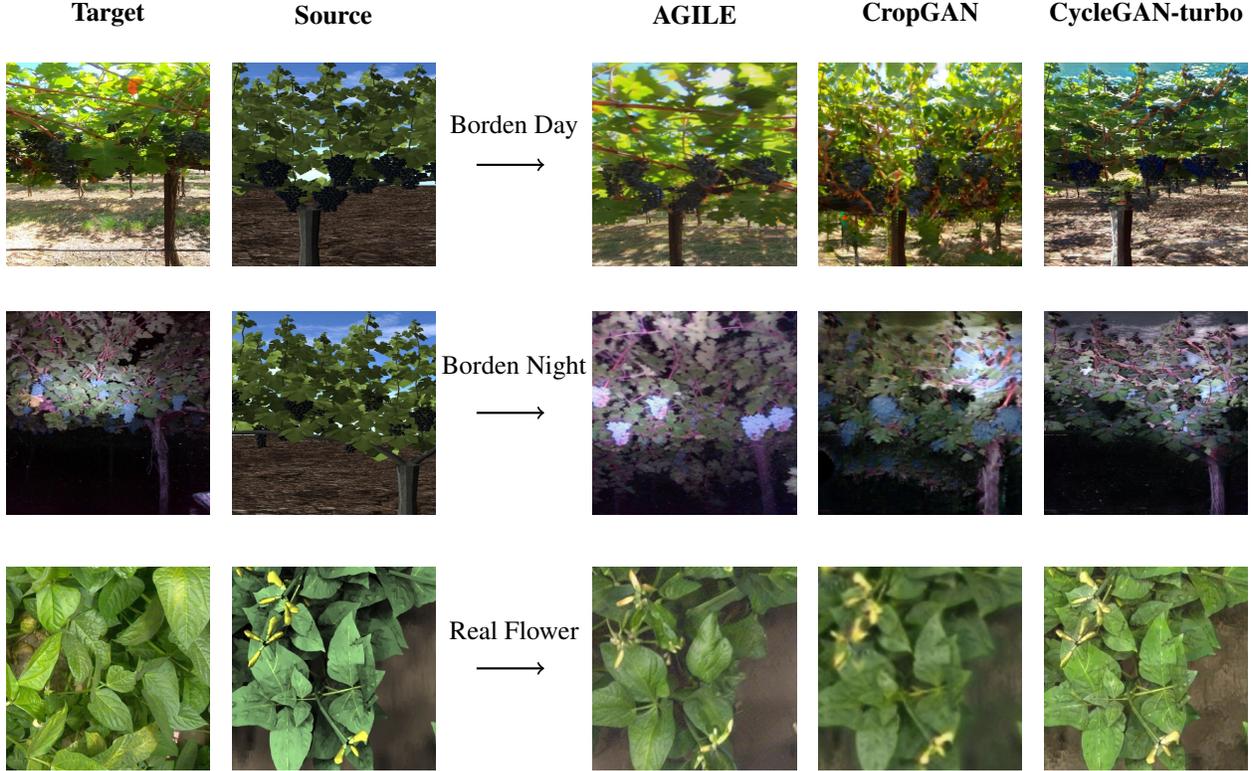
\begin{figure*}[h]
    \centering
    \begin{tikzpicture}
        \node at (-3.5,2) {\textbf{Target}};
        \node at (-0.5,2) {\textbf{Source}};
        \node at (4.3,2) {\textbf{AGILE}};
        \node at (7.3,2) {\textbf{CropGAN}};
        \node at (10.3,2) {\textbf{CycleGAN-turbo}};

        \node (img0) at (-3.5,0) {\includegraphics[width=2.7cm]{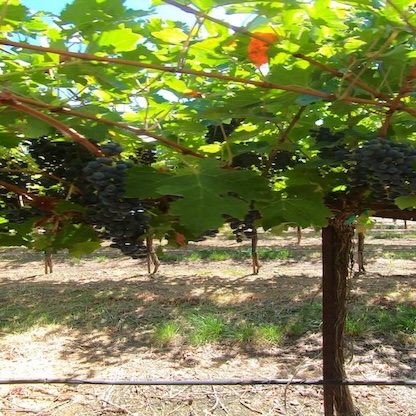}};
        \node (img1) at (-0.5,0) {\includegraphics[width=2.7cm]{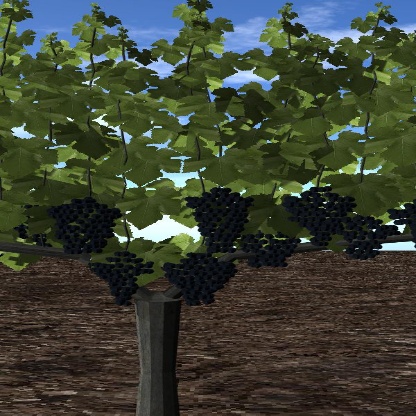}};
        \node (img2) at (4.3,0) {\includegraphics[width=2.7cm]{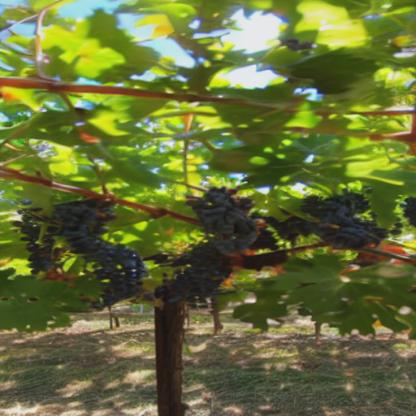}};
        \node (img3) at (7.3,0) {\includegraphics[width=2.7cm]{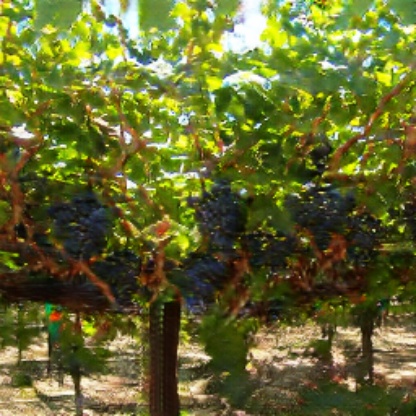}};
        \node (img4) at (10.3,0) {\includegraphics[width=2.7cm]{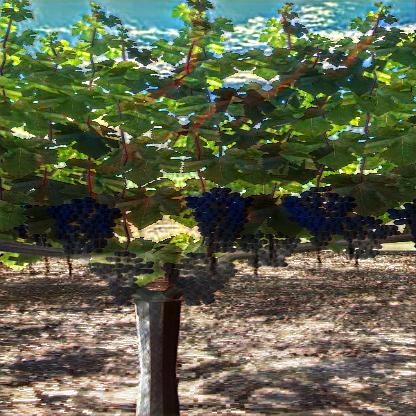}};

        \draw[->, thick] (1.4,0) -- (2.3,0);
        \node at (1.9,0.5) {Borden Day};

        \node (img5) at (-3.5,-3.3) {\includegraphics[width=2.7cm]{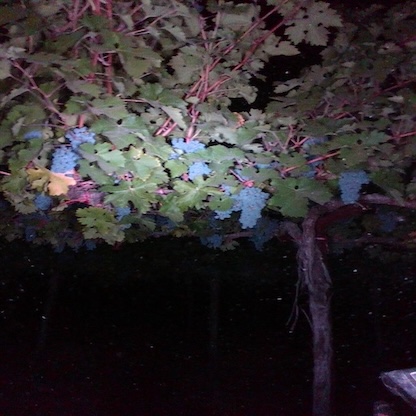}};
        \node (img6) at (-0.5,-3.3) {\includegraphics[width=2.7cm]{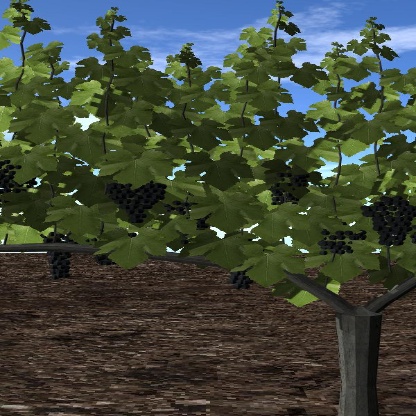}};
        \node (img7) at (4.3,-3.3) {\includegraphics[width=2.7cm]{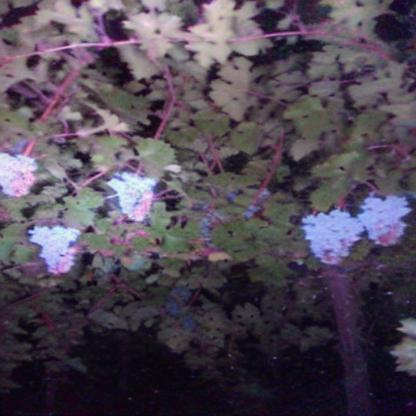}};
        \node (img8) at (7.3,-3.3) {\includegraphics[width=2.7cm]{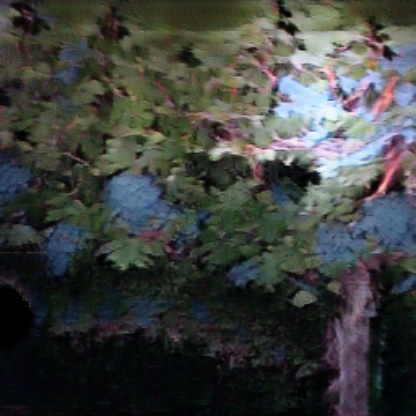}};
        \node (img9) at (10.3,-3.3) {\includegraphics[width=2.7cm]{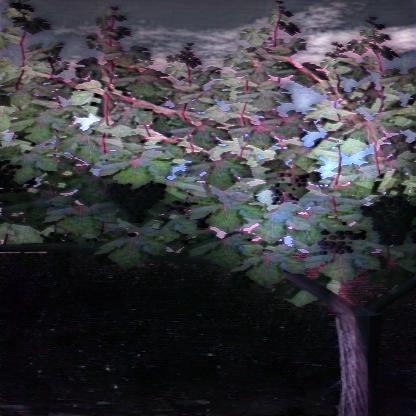}};

        \draw[->, thick] (1.4,-3.3) -- (2.3,-3.3);
        \node at (1.9,-2.7) {Borden Night};

        \node (img10) at (-3.5,-6.7) {\includegraphics[width=2.7cm]{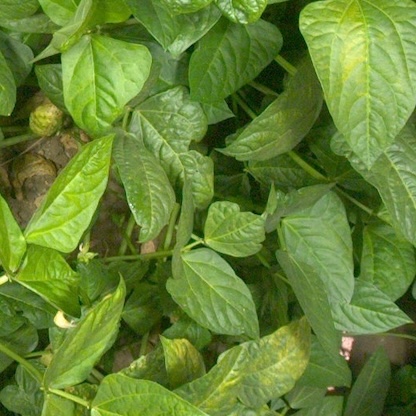}};
        \node (img11) at (-0.5,-6.7) {\includegraphics[width=2.7cm]{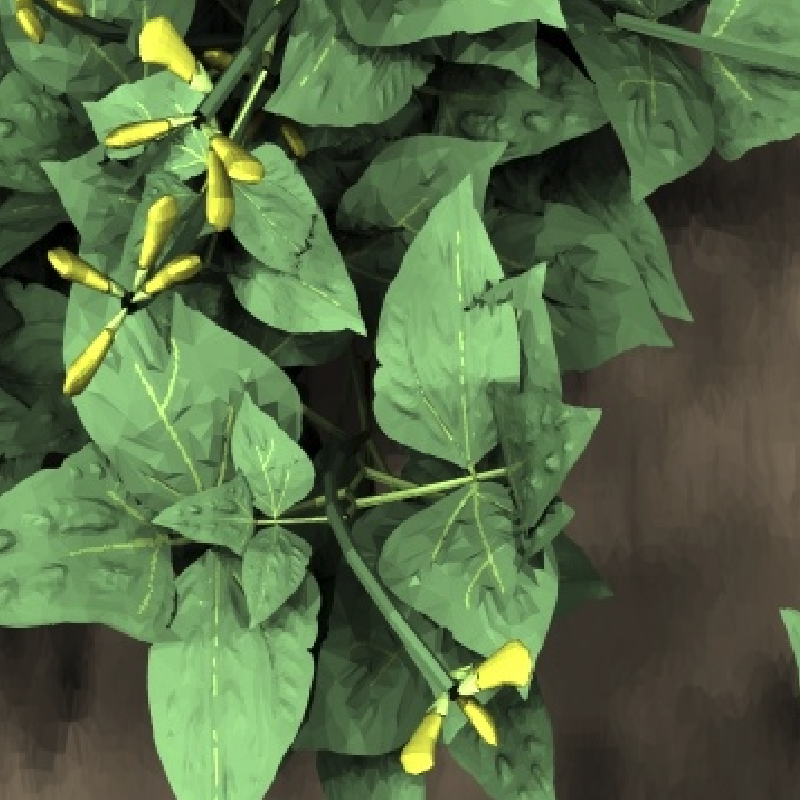}};
        \node (img12) at (4.3,-6.7) {\includegraphics[width=2.7cm]{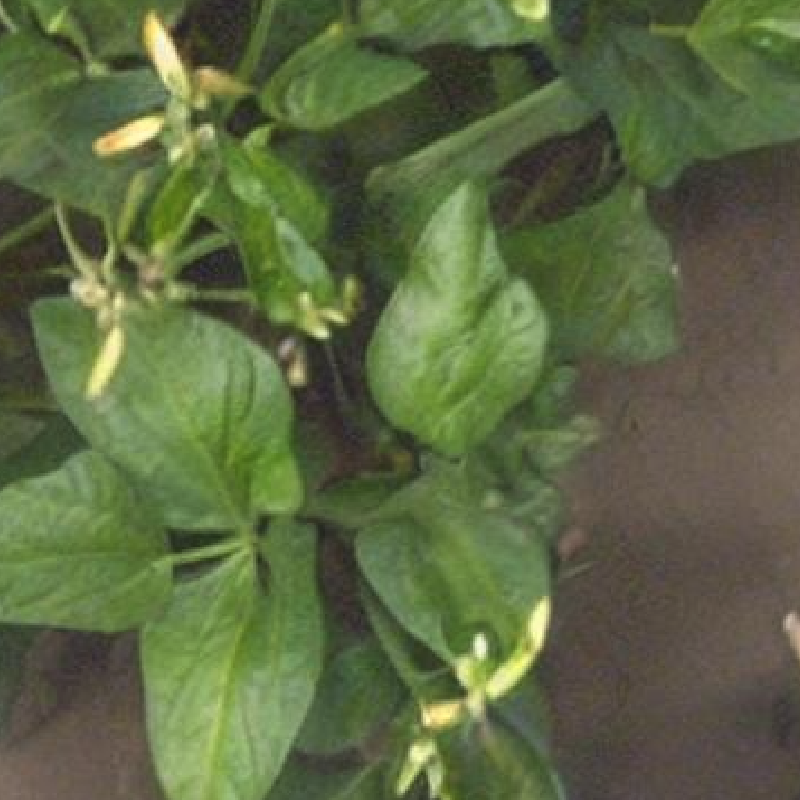}};
        \node (img13) at (7.3,-6.7) {\includegraphics[width=2.7cm]{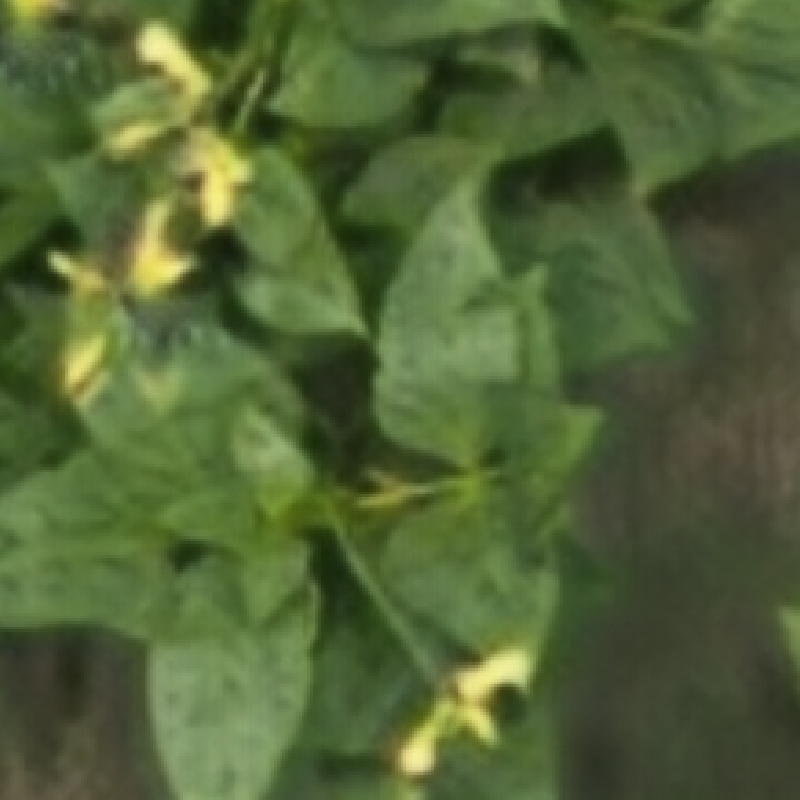}};
        \node (img14) at (10.3,-6.7) {\includegraphics[width=2.7cm]{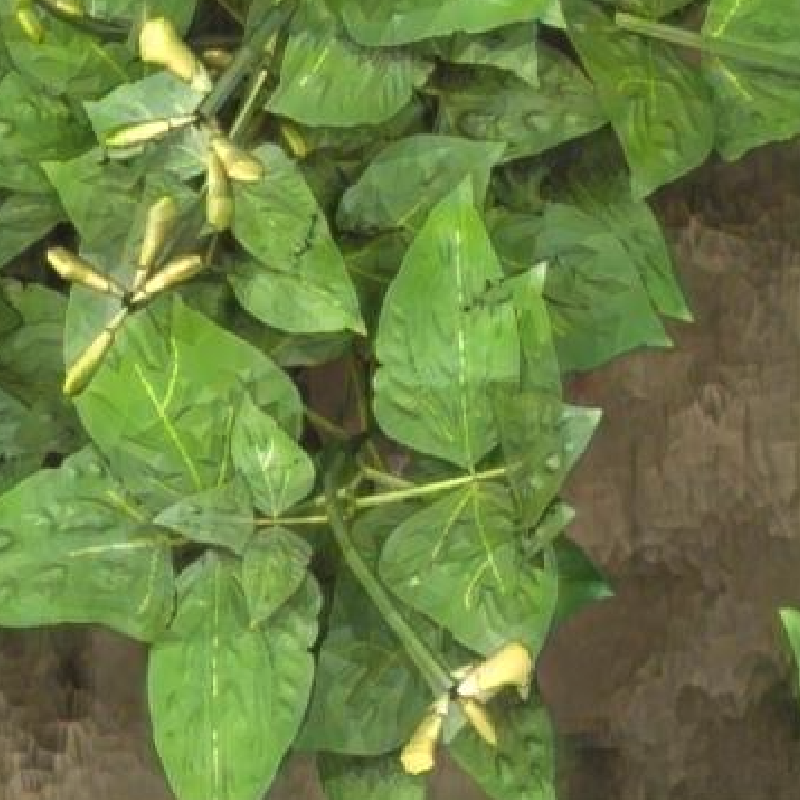}};

        \draw[->, thick] (1.4,-6.7) -- (2.3,-6.7);
        \node at (1.9,-6.2) {Real Flower};

    \end{tikzpicture}

    \caption{Generation results across translation tasks for our method (AGILE), CropGAN, and CycleGAN-turbo. The \textbf{Target} column represents the desired output domain for each translation task. \textbf{Top Row}: Synthetic Grape to Borden Day. \textbf{Middle Row:} Synthetic Grape to Borden Night. \textbf{Bottom Row:} Synthetic Flower to Real Flower.}
    \label{fig:compare_results}
\end{figure*}

\subsection{Datasets}

Each set of images is derived from AgML\footnote{https://github.com/Project-AgML/AgML}, a machine learning library for agricultural datasets. AgML provides labeled datasets of different plants in various domains. For example, we use a synthetically generated grape and flower dataset generated from Helios \cite{bailey_helios_2019}, a 3D Plant and Environment Biophysical Modeling Framework, as our source domain. Typically, synthetic images are treated as a source domain based on its capability to generate an infinite amount of labeled images. Additionally, plant modeling parameters can be tweaked to close the domain gap, in exchange for resources and compute time, to allow for improved generative results. As seen in Figure \ref{fig:datasets}, we train and evaluate our method on object detection tasks and constrain our translation within the same plant.

\subsection{Method}

We build the proposed method on top of ControlNet \cite{zhang2023addingconditionalcontroltexttoimage}, which enables conditional control of pretrained diffusion models by incorporating external control signals during the generation process. Our method first includes finding semantic correspondences between source and target images using pretrained diffusion models. In order to do that, we first optimize text embeddings using a query attention map generated from the labels of the source images. This allows us to have text-image correspondence without the need for paired images. Then, we use the optimized text embeddings to highlight regions of interest in the target domain. We further guide the attention maps to highlight the desired regions using the same set of query attention maps. This allows us to control the semantics of the target image through attention guidance during the denoising process of a diffusion-based model.

In diffusion models \cite{pmlr-v139-nichol21a}, during the forward process, an image \( I \) is encoded into its latent representation, \( x_0 \). Noise is gradually added to \( x_0 \) for \( T \) timesteps, making the data increasingly noisy until it becomes pure noise at \( t = T \). The training objective is to minimize the discrepancy between the model's predicted noise \( \epsilon_\theta(x_t, t) \) and the true noise \( \epsilon \). The model can be further conditioned on a text prompt \( y \) by providing an embedding \( e = \tau_\theta(y) \) using a text encoder \( \tau_\theta \):

\begin{equation}
    L = E_{x_0, \epsilon, t} \big[ \| \epsilon - \epsilon_\theta(x_t, t, e) \|_2^2 \big].
    \label{eq:training_objective}
\end{equation}

In the reverse process, using Denoising Diffusion Implicit Models (DDIM) \cite{song2021denoising}, a series of denoising steps, parameterized by \( \theta \), are applied to predict \( x_{t-1}\) given by:

\begin{equation}
    x_{t-1} = \sqrt{\bar{\alpha}_{t-1}} f_\theta(x_t, t) + \sqrt{1 - \bar{\alpha}_{t-1}} \epsilon_\theta(x_t, t),
    \label{eq:reverse_process}
\end{equation}

\noindent where \( \bar{\alpha}_{t-1} \) is the noise scaling factor, \( f_\theta(x_t, t) \) is the model's predicted denoised version of \( x_0 \), and \( \epsilon_\theta(x_t, t) \) is the model's prediction of the noise at step \( t \). The denoiser for Stable Diffusion based models is a transformer architecture \cite{NIPS2017_3f5ee243} utilizing a series of self-attention and cross-attention layers. For latent \( x_t \) at the \( t \)-th timestep, we can compute the attention maps by computing the query \( Q_l \) and key \( K_l \) attention map for each \( l \)-th layer of the decoder. The cross-attention for a given sample is defined as:

\small
\begin{equation}
    M_l(x_t \mid t, e, I_i) = \text{Attn}(Q_l, K_l) = \text{softmax}\left(\frac{Q_l K_l^\top}{\sqrt{d_k}}\right),
    \label{eq:cross_attention}
\end{equation} 

\noindent where \( d_k \) is the dimension of the key vectors. We first define our source and target model by fine-tuning the Stable Diffusion \cite{podell2023sdxlimprovinglatentdiffusion,rombach2022highresolutionimagesynthesislatent} model, SD1.5, until the input matches the output image, as seen in Section A and C from Figure \ref{fig:method}. During this fine-tuning process, we included geometric and color augmentations to the images concatenated during the denoising process, which was not originally implemented in ControlNet. Doing this prevents the model from overfitting with the training set.

\subsubsection{Text Optimization}

\begin{table*}[t]
    \centering
    \caption{Comparison of Fréchet Inception Distance (FID) and Average Precision (AP) metrics for various image translation methods across three domain translation tasks: Synthetic Grape to Borden Day, Synthetic Grape to Borden Night, and Synthetic Flower to Real Flower. Our method consistently outperforms existing approaches in both metrics. All methods were fine-tuned for each task. CropGAN was provided few-shot examples of labeled target images.}
    \label{tab:main_results}
    \small
    \renewcommand{\arraystretch}{1.2}
    \setlength{\tabcolsep}{1pt}
    \newlength{\metricwidth}
    \setlength{\metricwidth}{0.12\textwidth}

    \begin{tabular}{@{\hspace{4pt}}l*{6}{>{\centering\arraybackslash}m{\metricwidth}}@{\hspace{4pt}}}
        \hline
        \multirow{3}{*}[-1ex]{\textbf{Method}} & 
        \multicolumn{2}{c}{\makecell{\textbf{Syn. Grape}\\to\\\textbf{Borden Day}}} & 
        \multicolumn{2}{c}{\makecell{\textbf{Syn. Grape}\\to\\\textbf{Borden Night}}} & 
        \multicolumn{2}{c}{\makecell{\textbf{Syn. Flower}\\to\\\textbf{Real Flower}}} \\[1ex]
        \cline{2-7}
        & \small\textbf{FID}$\downarrow$ & \makecell{\small\textbf{AP}$\uparrow$} & 
        \small\textbf{FID}$\downarrow$ & \makecell{\small\textbf{AP}$\uparrow$} & 
        \small\textbf{FID}$\downarrow$ & \makecell{\small\textbf{AP}$\uparrow$} \\
        \hline
        Synthetic Only & 265.96 & 0.12 & 363.87 & 0.00 & 175.94 & 0.11 \\
        \rowcolor{white} {\small \textcolor{gray}{\textit{Real Only}}} & {\small \textcolor{gray}{\textit{92.89}}} & {\small \textcolor{gray}{\textit{0.51}}} & {\small \textcolor{gray}{\textit{129.27}}} & {\small \textcolor{gray}{\textit{0.69}}} & {\small \textcolor{gray}{\textit{81.93}}} & {\small \textcolor{gray}{\textit{0.65}}} \\
        \hdashline
        CropGAN\cite{fei_enlisting_2021} & 131.18 & 0.28 & 192.12 & 0.17 & 202.96 & 0.24 \\
        CycleGAN-turbo\cite{parmar_one-step_2024} & 146.64 & 0.09 & 276.94 & 0.00 & \textbf{154.21} & 0.33 \\
        ControlNet\cite{zhang2023addingconditionalcontroltexttoimage} & 140.60 & 0.09 & 198.90 & 0.00 & 183.53 & 0.26 \\
        \hdashline
        Ours & \textbf{123.08} & \textbf{0.33} & \textbf{187.18} & \textbf{0.32} & 156.44 & \textbf{0.35} \\
        \hline
    \end{tabular}
\end{table*}

Images collected in the field do not come with a paired text description. Therefore, we cannot collect semantic correspondences between images using paired data. To address this, we propose to optimize provided text embeddings using existing labeled images for improved semantic knowledge, as summarized in Part B of Figure~\ref{fig:method}. The labeled images, the source domain, must contain the object of interest that makes semantic correspondence between the source and target domain possible. A query attention map, \( M_s \), is created for each labeled image containing multiple Gaussian markers centered within the labeled bounding box. Each Gaussian marker, \( G(x, y) \), is defined as:

\begin{equation}
    G(x, y) = \exp\left(-\frac{(x - x_c)^2}{2\sigma_x^2} - \frac{(y - y_c)^2}{2\sigma_y^2}\right),
    \label{eq:gaussian_map}
\end{equation}

\noindent where \( (x_c, y_c) \) is the center of the Gaussian marker and \( \sigma_x^2 \) and \( \sigma_y^2 \) are the variances. The variance can be scaled by the dimensions of the bounding boxes. After creating the query attention map for each labeled image, we condition the stable diffusion model with the text embedding \( e \) and optimize the text embedding at a specific timestep.
As the Gaussian regions represent the desired region of focus, we can optimize the embedding \( e \) that reproduces the desired attention map. We extracted the cross-attention map from timestep 30 and optimize the first token only using MSE loss:

\begin{equation}
    e^* = \arg \min_e \frac{1}{N} \sum_{i=1}^N \|M_l(x_t \mid t, e, I_i) - M_s(I_i)\|_2^2.
    \label{eq:optimize_embedding}
\end{equation}

The attention response from each layer in the decoding stage exhibits different levels of semantic knowledge \cite{mokady_null-text_2022}. Therefore, to summarize these responses, we average the attention maps for token 1 across all the decoding layers and attention heads. At the end of each optimization step, we obtain an optimized text embedding \( e^* \) that highlights the desired regions of interest in the source domain. Once \( e* \) is obtained from \eqref{eq:optimize_embedding}, it is used as input for subsequent optimization steps using \( M_l(x_t \mid t, e*, I_i) \). We have found that at least five images are needed to get the desired optimized embedding.

\subsubsection{Attention Guidance}

The following steps are applied after defining the target model as shown in Part C in Figure~\ref{fig:method}. Our approach builds on ControlNet by concatenating the source and synthetic images during the denoising process to preserve semantic consistency, while skip connections help maintain fine details \cite{parmar_one-step_2024}. In contrast to text optimization, we target the first three cross-attention layers of the decoding block, which are closest to the high-level latent feature space. Within each of these layers, we aggregate the attention maps by averaging over all attention heads and spatial dimensions to obtain a single representative map per token, \( M_{l,token} \). Since we employ single-word text embeddings, we designate the first token as the ``object" token and treat the remaining tokens as ``background" tokens. This allows us to apply different scaling weights to object and background tokens. Prior to scaling, we compute the mean and standard deviation of each map, \( M_{l,token} \), for the object and background token separately, and normalize the query map based on these statistics. As a result, we get normalized representative maps per token for each selected decoding layer, \( \tilde{M}_{l,token} \). Since each layer produces attention maps at different resolutions, the representative maps are interpolated to match the dimensions of the query map, ensuring consistency during the attention guidance process.

We add the final query map \( \tilde{M}_s(I_i) \) with the final attention map \( \tilde{M}_{l,token} \) for each token index as shown:

\begin{equation}
    M^{edit}_{l,token} = \beta*\tilde{M}_{l,token}(x_t | t, e^*, I_i) + \tilde{M}_s(I_i),
    \label{eq:attention_edit}
\end{equation}

\noindent where \( \beta \) is the scaling factor that can be different for object or background tokens. This editing approach can be stopped early to prevent harsh contrast differences with the target domain, as seen in Figure~\ref{fig:timesteps}, though this can change depending on the dataset used. 
\section{Experiments and Ablation Study}
\label{sec:results} 

Our method leverages synthetic images and their corresponding labels to enforce semantic constraints during image-to-image translation. In our framework, the real images are treated as the target domain, but their labels are withheld during training to simulate an unpaired setting. To evaluate our approach, we compute the Fréchet Inception Distance (FID) \cite{heusel_fid_2018} between the generated images and the target domain. For comparison, we also calculate the FID for the Synthetic Only and Real Only baselines by measuring the distance between their respective training sets and the target test set.

To demonstrate the practical utility of our generated images, we use them as training data for a Faster R-CNN \cite{ren_faster_2016} object detection model. Specifically, we train the detection model on 
 the synthetic, real and generated images annotated with bounding boxes corresponding to grape clusters or flowers. After training, we test the model on real images from the target domain, comparing the predicted bounding boxes to the ground truth bounding boxes from the real dataset. We quantify detection performance using Average Precision (AP). Finally, we compare our results against recent unpaired image-to-image translation methods to highlight the effectiveness of our approach.

\begin{table}[t]
    \centering
    \caption{Ablation study results summarizing the importance of text optimization prior to attention guidance.}
    \label{tab:ablation}
    \small
    \renewcommand{\arraystretch}{1.2}
    \setlength{\tabcolsep}{1.5pt} 
    \newlength{\tightwidth}
    \setlength{\tightwidth}{0.23\columnwidth} 

    \begin{tabular}{@{\hspace{1.5pt}}l*{3}{>{\centering\arraybackslash}p{\tightwidth}}@{\hspace{2pt}}}
        \hline
        \multirow{2}{*}[-1ex]{\textbf{Method}} & 
        \makecell{\textbf{Syn. Grape}\\to\\\textbf{Borden Day}} & 
        \makecell{\textbf{Syn. Grape}\\to\\\textbf{Borden Night}} & 
        \makecell{\textbf{Syn. Flower}\\to\\\textbf{Real Flower}} \\[1ex]
        \cline{2-4}
        & \small\textbf{AP$\uparrow$} & \small\textbf{AP$\uparrow$} & \small\textbf{AP$\uparrow$} \\
        \hline
        No Text Optim. & 0.10 & 0.01 & 0.13 \\
        No Guidance & 0.14 & 0.00 & 0.25 \\
        \hdashline
        Ours & \textbf{0.33} & \textbf{0.32} & \textbf{0.35} \\ 
        \hline
    \end{tabular}
\end{table}

\subsection{Quantitative Results}
We evaluate our approach against CropGAN \cite{fei_enlisting_2021}, CycleGAN-turbo \cite{parmar2023zeroshotimagetoimagetranslation}, and ControlNet \cite{zhang2023addingconditionalcontroltexttoimage}. Table~\ref{tab:main_results} presents a quantitative comparison based on FID between the generated images and the target domain, as well as the AP when performing object detection on the target domain.

Our results demonstrate that AGILE outperforms the baseline models across all datasets in terms of AP. For example, when translating Synthetic Grape to Borden Day, AGILE achieves an AP of 0.33, which is a significant improvement compared to CropGAN (0.28) and CycleGAN-turbo (0.09). Similarly, for the Synthetic Grape to Borden Night task, AGILE achieves an AP of 0.32, surpassing both CropGAN (0.17) and CycleGAN-turbo (0.00). Additionally, when translating Synthetic Flower to Real Flower, AGILE achieves an AP of 0.35, outperforming CropGAN (0.24) and slightly surpassing CycleGAN-turbo (0.33).

While AGILE excels in AP performance across all tasks, CycleGAN-turbo achieves a marginally better FID score for the Synthetic Flower to Real Flower translation task. Despite this, AGILE demonstrates superior performance in maintaining object semantics and producing high-quality images across the other domains. Furthermore, our approach surpasses CropGAN, a few-shot learning method, across all datasets, despite CropGAN being provided with some of the labeled target images.

We also conducted an ablation study to assess the impact of text optimization on attention guidance. As shown in Table~\ref{tab:ablation}, when text optimization is omitted, attention guidance adversely affects the generation process compared to both the base ControlNet method and the approach using only text optimization.

\begin{figure}[t]
    \centering
    \includegraphics[width=0.7\linewidth]{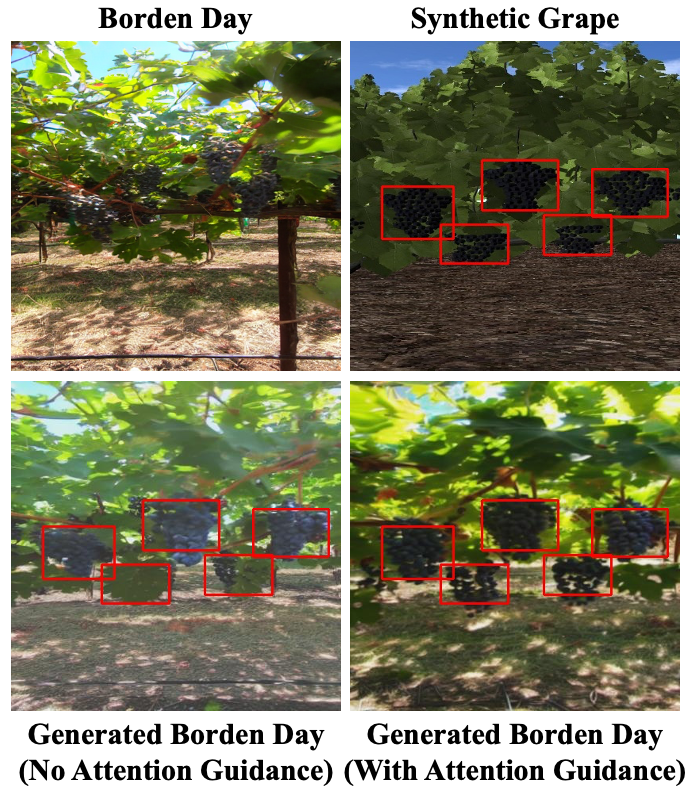}
    \caption{Without attention guidance, the generated image fails to translate the labeled objects. The background and object color of the generated Borden Day image is inconsistent with the target domain.}
    \label{fig:ablation_2}
\end{figure}

\subsection{Qualitative Comparisons}
Figure~\ref{fig:compare_results} presents a visual comparison of the images generated by the different approaches. Our method preserves realism while ensuring semantic consistency, whereas CropGAN and CycleGAN-turbo fall short. For example, CropGAN struggles to distinguish between the sky and grape regions in the Synthetic Grape to Borden Night translation, and CycleGAN-turbo fails to maintain proper color consistency with the target domain. Overall, our approach effectively preserves the color, shape, and texture of both object and background compared to the other methods. For instance, a grape in the synthetic image could look very different in the target image. But our approach aims to maintain the shape despite differences between domains. However, it does encounter challenges when transferring very small objects that provide minimal signal, as seen in the flower translation task.

\begin{figure}[t]
    \centering
    \includegraphics[width=1.0\linewidth]{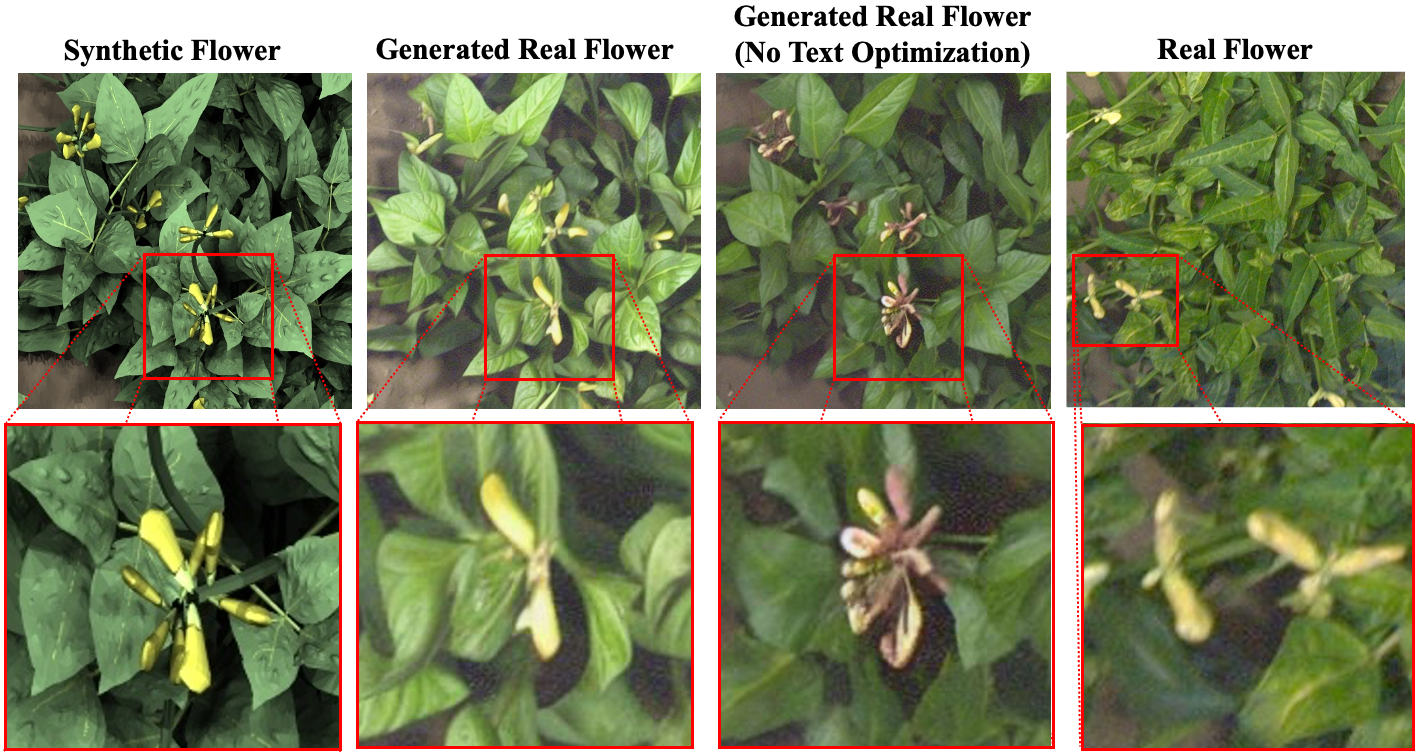}
    \caption{Without text optimization, attention guidance fails to accurately generate the correct shape and color of the flower in the real image. Despite differences between the synthetic and real flowers, our method successfully preserves the flower’s shape within the target domain, demonstrating improved semantic consistency.}
    \label{fig:ablation}
\end{figure}

In our ablation study, we attempted attention guidance using a text embedding that is not optimized to have object knowledge, as seen in Figure~\ref{fig:ablation}. As a result, the method without text optimization failed to preserve the object's shape and color in the target domain. Additionally, we observed an inconsistency in the background color between the full proposed method and the variant without text optimization, though this discrepancy may be unrelated. This proves that text optimization is a necessary preliminary step prior to attention guidance. Figure~\ref{fig:ablation_2} shows that without attention guidance, the generated image fails to translate the labeled objects and struggles to maintain object color consistency with the target domain.
\section{Discussion}
\label{sec:discussion}

The timing and location for optimizing text embeddings and guiding attention maps depend on the specific timestep chosen for applying edits or optimizations during the denoising process, as well as the decoding layers targeted for modification. We optimize at $t=30$ while noise is still present which ensures that the optimized text embedding maintains robust against variability in images. Furthermore, we edit only the first three layers to avoid object distortion or inconsistent colors in the final image. Since attention guidance involves manipulation the attention scores, excessive adjustments could lead to unwanted artifacts or background inconsistency. 

Additionally, we found that different layers in the decoding block attend to different areas of the image. For example, early layers focus on interpreting the foreground or objects defined by the text embedding, while later layers contribute to background refinement and overall image coherence. Overall, there are two key axes to consider for successful text optimization and attention guidance: the denoising timestep and the decoding layers.

The scaling of the attention maps using $\beta$ from Equation~\ref{eq:attention_edit} enhances signals for particularly small objects, ensuring they receive adequate focus. Multiplying this scalar on background tokens improves semantic consistency, even though these tokens are not explicitly optimized to contain the specified object knowledge. ControlNet provides a mechanism to preserve the semantics of the source image by scaling the concatenated feature map with a factor known as the control scale during the decoding process. However, we observed that increasing this value beyond one degrades image quality. Conversely, decreasing this value when there are large domain gaps—such as between synthetic grape images and Borden night scenes—yields better results by preventing the output from overly adhering to the source domain.
\section{Limitations}
\label{sec:limitations}

Our method faces challenges when translating small objects, as object guidance via attention scores is constrained by the resolution of the attention map. If an object is too small, the corresponding signal may be too weak to be effectively captured, leading to reduced visibility in the final output. Additionally, our method also struggles with unseen examples. For instance, if an object is found in the synthetic image but is not found in the real image, the model will try to fill in the gaps with plausible details based on its learned prior, ensuring coherence with the training distribution. This means that the model synthesizes missing features in a way that aligns with its learned representations, potentially leading to hallucinations if the object is significantly different from those seen during training. 

A potential solution for improving small object translation and overall image quality is enabling guidance for multiple objects, which requires incorporating multi-object knowledge within the text embedding. Our proposed method currently relies on labels for a single class, but extending the approach to optimize for multiple classes could make multi-object guidance feasible.

Despite applying these image translation methods, domain gaps still exist between the generated target images and the ground truth target images. Our visual analysis revealed that the most influential factors contributing to this gap are perspective, brightness, and plant type. These parameters are not explicitly accounted for during the diffusion process, as they are not defined on a per-image basis.
\section{Conclusion and Future Work}
\label{sec:conclusion}

In this paper, we proposed AGILE (Attention-Guided Image and Label Translation for Efficient Cross-Domain Plant Trait Identification), a diffusion-based framework aimed at improving semantic consistency for cross-domain image translation tasks relevant to plant trait identification. By leveraging pretrained diffusion models, optimized text embeddings, and attention guidance during the denoising process, AGILE effectively maintains object structure and semantics even when there are significant domain gaps.

Our experimental results demonstrate that AGILE consistently outperforms existing image translation methods across multiple datasets. The quantitative results highlight improvements in object detection performance, while the qualitative comparisons show enhanced realism and consistency in generated images. Furthermore, ablation studies validate the importance of text optimization and attention guidance, emphasizing their complementary roles in preserving semantic alignment between source and target domains.

However, our approach still faces challenges when translating small objects or generalizing to unseen examples. Additionally, existing domain gaps such as perspective, brightness, and plant type are not fully addressed during the diffusion process. While the domain-translated images generated by AGILE may not completely surpass the performance of using real, labeled target images, they can still provide valuable information for pretraining a backbone model. This pretraining can enhance the model’s ability to extract relevant features, which can then be further fine-tuned using a limited set of labeled images from the target domain.

Future work will focus on extending our method to incorporate multi-object guidance through enhanced text embeddings and improving robustness to various domain gaps. Furthermore, we will explore more efficient optimization techniques to enhance performance and generalization across diverse agricultural datasets.
{
    \small
    \bibliographystyle{ieeenat_fullname}
    \bibliography{main}
}


\end{document}